# Implementation of Fruits Recognition Classifier using Convolutional Neural Network Algorithm for Observation of Accuracies for Various Hidden Layers

Shadman Sakib[1], Zahidun Ashrafi[2], Md. Abu Bakr Sidique[3]
[1,3]Department of EEE, International University of Business Agriculture and Technology, Dhaka 1230, Bangladesh
[2]Department of Statistics, Rajshahi College, Rajshahi 6100, Bangladesh
sakibshadman15@gmail.com[1], zimzahidun@gmail.com[2], absiddique@iubat.edu[3]

*Abstract*—Fruit recognition using Deep Convolutional Neural Network (CNN) is one of the most promising applications in computer vision. In recent times, deep learning based classifications are making it possible to recognize fruits from images. However, fruit recognition is still a problem for the stacked fruits on weighing scale because of the complexity and similarity. In this paper, a fruit recognition system using CNN is proposed. The proposed method uses deep learning techniques for the classification. We have used Fruits-360 dataset for the evaluation purpose. From the dataset, we have established a dataset which contains 17,823 images from 25 different categories. The images are divided into training and test dataset. Moreover, for the classification accuracies, we have used various combinations of hidden layer and epochs for different cases and made a comparison between them. The overall performance losses of the network for different cases also observed. Finally, we have achieved the best test accuracy of 100% and a training accuracy of 99.79%.

I. INTRODUCTION

With the lively improvement of our human society, additional attention has been paid to the superiority of our life, particularly the food we eat. Over the last few years, computer visions have been widely used in fruit recognition methods. In the field of image recognition and classification, Deep Neural Network (DNN) is used to identify fruits from images. DNN performs better than other machine learning algorithms. Convolutional Neural Networks (CNNs) are classified as a deep learning algorithm. In deep learning, CNN [1, 2] are the most commonly used type of Artificial Neural Networks (ANNs). It is being used several visual recognition analyzing which includes video and image recognition [3], face recognition [4], handwritten digit recognition [5], and fruit recognition [6] etc. The accuracies in these fields including fruit recognition using CNN have reached human-level perfection. Mammalian visual systems' biological model is the one by which the architecture of the CNN is inspired. It was found by D. H. Hubel et al. in 1062 that the cells in the cat's visual cortex are refined to a minute area of the visual field which is recognized as the receptive field [7]. In 1980, the neocognitron [8] introduced by Fukushima was the pattern recognition model inspired by the work of D. H. Hubel et al. [9] was the first computer vision. However, CNN is categorized by a network architecture which consists of convolution and pooling layers to extract and combine high-level features from 2D input.

CNN has a very similar architecture as ANN. There are several neurons in each layer in ANN. Hence, the weighted sum of all the neurons of a layer becomes the input of a neuron of the next layer adding a biased value. In CNN the layer has three dimensions. Here all the neurons are not fully connected instead they are connected to the local receptive field. A cost function is generated in order to train the network. It compares the network's output with the desired output.

Accurate and efficient fruit recognition is of great importance in the field of robotic harvesting and yield mapping. An ideal fruit recognition system is accurate that can be trained on an easily available dataset, shows real-time predictions and acclimates various types of fruits. Therefore, in our research, we implement a fruit recognition classifier using CNN. The input image is taken as 100×100 pixels of RGB image. For the networks best performance, we used various combinations of hidden layers for five cases and observe the accuracies. The final experiment result shows the much-improved fruit recognition rate. The mathematical model of the network is executed in python with tensorflow.

II. PREVIOUS WORK

A number of factors made fruit recognition a challenging task which includes fruits that arise in scenes of fluctuating brightness, obstructed by other objects, sharpen edge, texture, reflectance properties etc. Many kinds of research exist to help fruit recognition challenges. Fruit recognition can be considered as an image segmentation problem. Several works are available in the literature addressing the problem of fruit recognition as an image segmentation problem. Wang et al. [10] established a system that detects apples based on their color. They surveyed the issue of apple detection for yield prediction. Hung et al. [11] proposed a five-class segmentation method for almond segmentation using conditional random fields. This method learned features using a Sparse Autoencoder (SAE). Later, these features were used within the CRF framework and shown remarkable segmentation performance. A novel approach for detecting fruit using the deep convolutional neural network is presented in paper [6]. In this paper, the author adapts a Faster Region-based CNN through transfer learning. They trained the model using RGB and NIR (Near-Infrared) images. The combination of RGB and NIR discovers the early and late fusion methods. Fruit recognition method based on deep fruit system [6] achieved a remarkable milestone in the development of deep learning approaches for fruit detection. In another research [12], where the neural networks trained by two backpropagation algorithms on images of Apple Gala variety trees in order to calculate the yield for the forthcoming seasons. Furthermore, detection in relation with the angle of the camera [13], Scale Invariant Feature Transform (SIFT), improved ChanVese level-set model [14] are also used in the literature for the fruit recognition.

### III. OVERALL ARCHITECTURE OF THE PROPOSED CNN MODEL

The network consists of two convolutional layers, each of them followed by pooling layers, and two fully connected layers shown in figure 1. The input layer of the network contains 30,000 neurons as input data, representing the standard RGB image of size 100×100 pixel. The first hidden layer is the convolutional layer 1 which has 64 filters with a kernel of size 3×3 pixels and Rectified Linear Units (ReLU) as an activation function. The second convolutional layer is the convolutional layer 2 where 64 filters with the kernel size of 3×3 pixels and ReLU were employed as on the first convolutional layer. The convolutional layer is used for the feature extraction from input data. Also performs convolution operation to small localized areas by convolving a filter with the previous layer. The kernel size determines the locality of the filters. A kernel initializer named Lecun uniform is used along with every hidden layer for initializing the weights. ReLU is used as an activation function to enhance the performance at the end of all convolutional layers and fully connected layers. In pooling layers 1 and 2, where max pooling is used with a pool size of 2×2 along with a stride length of 2. The padding used here is the same padding which means that the output and input feature maps has the same spatial dimensions. Pooling layers minimize the spatial size of the output and controls overfitting. The stride defines how the convolution operation works with a kernel when the larger sizes of an image and complex kernels are used. A regularization layer dropout with a probability of 0.25 is used next to the pooling layer 2 where it randomly switches off 25% of the neurons in the layer during training hence reducing overfitting as well as to improve the performance of the network by making it more robust. This causes the network to become capable of better generalization and less compelling to overfit the training data.

$$\sigma(z) = (1+\exp(-z))-1 \qquad (1)$$

After the dropout, a flatten layer is used which converts the 2D filter matrix into 1D feature vector before entering into the fully connected layers. The next hidden layer is the fully connected layer 1 consists of 500 neurons with a dropout of 0.5 and ReLU. Again a dropout with a probability of 0.5 is used between the output layer and the last hidden layer. Finally, the fully connected layer 2 or the output layer contains 25 neurons where softmax classifier activation is used to predice the output of the model and represents 25 different classes of fruits.

$$\sigma(z)_j = \frac{e^{z_j}}{\sum_{k=1}^{k} e^{z_k}} \quad \text{for j=1, ...k} \qquad (2)$$

In deep neural network, optimizer plays a vital role and helps to decrease or increase the error function of the model. For training our network we used Adam [15] optimizer. Adam stands for Adaptive Moment Estimation which computes adaptive learning rates for hyper-parameter. The Adam optimization algorithm is straightforward to implement, requires little memory, appropriate for problems with sparse gradients and computationally effective. Moreover, it is broadly used in many deep learning applications like computer vision and natural language processing. For our CNN model, we used Adam optimizer with a learning rate of 0.002, batch size of 15 and 15 epochs. For reducing the model's training error a small learning rate is very important. For estimating the moments, Adam deploys exponentially decaying averages and computed on the gradient as:

$$n_t = \beta_1 * n_{t-1} + (1-\beta_1)*g_t \qquad (3)$$

$$s_t = \beta_2 * s_{t-1} + (1-\beta_2)*g_t^2 \qquad (4)$$

Where $n_t$ and $s_t$ are the first and second moments, $g_t$ is the gradient, $\beta_1$ and $\beta_1$ are the hyper-parameters.

For performing the weight update:

$$w_t = w_{t-1} - \eta * \frac{\hat{n}}{\sqrt{\hat{s}}+\epsilon} \qquad (5)$$

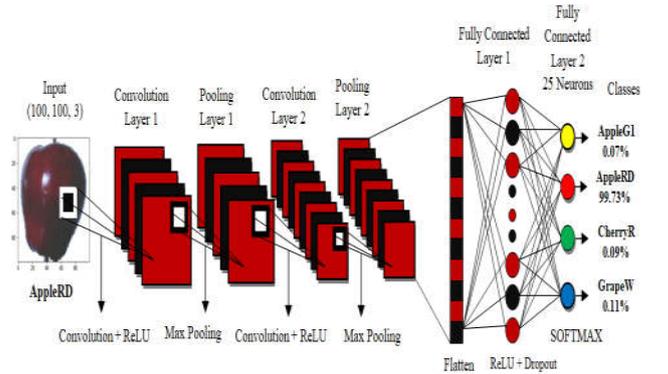

Fig. 1. Schematic of the architecture of a convolutional neural network

To compute the model performance, error rate need to be calculated. We have used categorical cross-entropy cost function as a loss function.

$$L_i = -\sum_j t_{i,j} \log(p_{i,j}) \qquad (6)$$

Cost function is defined as [16]:

$$C(w,b) = \frac{1}{2n} \sum_x \left[y(x)-a^2\right]^2 \qquad (7)$$

Where w is the cumulation of weights in the network, b is all the biases, n is the total number of training inputs and a is the actual output. C(w,b) is non-negative as all the terms in the sum is non-negative. To reduce the cost C(w,b) to a smaller degree as a function of weight and biases, the training algorithm has to find a set of weight and biases which cause the cost to become as small as possible. An automatic learning rate reduction technique [17] is used to make the optimizer coincide faster and closer to the global. Figure 2 shows the graphical illustration for achieving the global and local cost minimum.

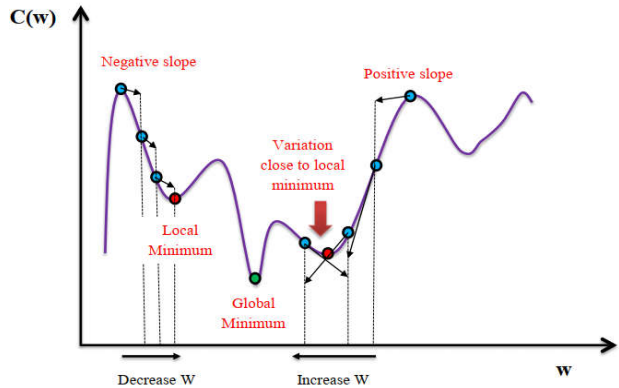

Fig. 2. Graphical illustration of Cost against weight

When the learning rate is low, it takes more time to reach into the global minimum and the training cannot coincide nor differ, when the learning rate is higher.

## IV. DATASET

For training and testing, all the images were selected from the fruits-360 dataset which is publicly available on Github and Kaggle [17]. The dataset contains 65,429 different fruit images of 95 categories. The fruit images were reaped by recording the fruits while they are revolved by a motor and then producing frames. A white paper is placed behind the fruits as a background. Due to the disparity in the lighting a flood fill type algorithm was developed which extract the fruit from the background. After removing the background all the fruits were scaled down to 100×100 pixels of standard RGB fruit images.

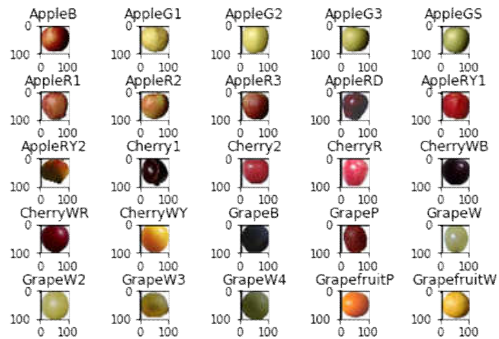

Fig. 3. Sample of different fruit images from fruits-360 dataset

From the fruits-360 dataset, we picked 17,823 images from 25 different categories. Among them to create the training set we used 14,258 images (80%) and the rest 3,565 images (20%) used for testing the model. Figure 3 shows the 25 different categories of fruits we used for finding the classification accuracy. The network is trained for 15 epochs with a batch size of 15.

## V. EXPERIMENTAL RESULTS

In this paper, we have applied a convolutional neural network on the fruits-360 dataset in order to find the better classification performance of the network. For determining the overall classification accuracies, we have taken five cases where we applied different combinations of hidden layers (convolution and pool) for 15 epochs with a mini batch size of 15 and calculated the accuracies on the test and training set. The accuracies were obtained using tensorflow library in python and the simulation has done using MATLAB software.

Figure 4 compares the overall training and test accuracies for various combinations of hidden layers. The highest classification accuracy on the test images was found 100% attained by case 4 (Conv1, pool1, Conv2, pool2 with dropout). This shows the greatest performance of the network on the test set which consecutively occurred at 11 up to 15 epochs. The next highest test accuracy has been obtained for case 1, case 2, and case 3 (99.94%) at epoch 15, epoch 4, and epoch 15 respectively. On the other hand, case 1 (Conv1, pool1, Conv2, pool2 without dropout) obtained the highest recognition accuracy on the training images which is 99.79% at epoch 15. Training the CNN by various combinations of hidden layers and increasing the number of epochs resulted in the highest test accuracy than the training accuracy.

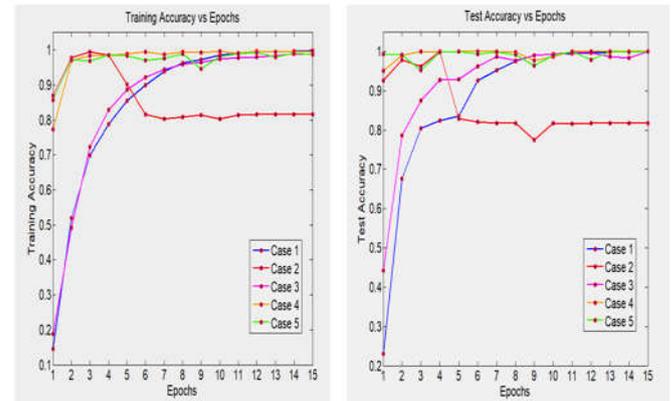

Fig. 4. Training and test accuracy curves for different epoch accompanied by various combinations of convolution and pool

The difference between training and test recognition accuracies is the highest (29.39%) for case 3. Similarly, case 4 and case 5 also has a big difference between training and test recognition accuracies. This is because of the overfitting of the model to the training data. However, the subsequent best training accuracy was found 99.73% at epoch 10.

Figure 5 shows the losses for different epoch for different combinations of convolution and pool. Different responses were found in different combination.

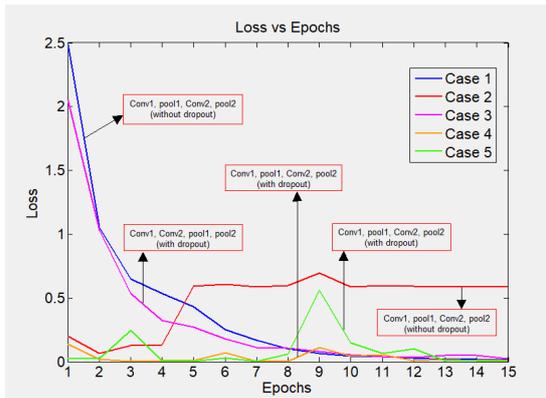

Fig. 5. Loss curves for CNN for different epoch accompanied by various combinations of convolution and pool

For the first case convolution and pool were used alternately without dropout. The loss curve responded in a downward direction. The loss was the most for the early 4 epochs and gradually decreased up to 15 epochs. The loss soothed approximately at 0.0193 from 13 up to 15 epochs. Again, for case 2 where convolution and pool were used periodically without any dropout shows the highest performance loss of 0.5881. The loss was at a minimum at the beginning and steadied twice between 5 to 8 epochs and 10 to 15 epochs. Similarly, when two convolutions and two pools were used with dropout (case 3), the loss curve shows the identical performance similar to case 1 except that for case 3 the loss curve was steepest between 1 to 4 epochs. For this case, the overall test loss was 0.0222.

Concurrently, the fourth case shows the less performance loss (0.0078) than other cases apart from infinitesimal fluctuations in the interval from 8 to 9 epochs. On the contrary, the fifth case where convolution and pool were used in a repeated order with dropout shows the finest performance of CNN classifier. For this case, the peak loss was approximately 0.0291 and increased little from 2 to 3 epochs. Afterward, the loss smoothly deteriorated to below 0.0021 and increased intensely again from 8 to 9 epochs. A few more fluctuations occurred and finally, the loss was 0.0032 becomes steady near about 15 epochs which is the smallest loss of the performance. It is shown that CNN behaves differently for different combinations of layers. However, from our result, we have found that the classification performance gets better if the convolution and pool are used in a repeated order with dropout.

## VI. CONCLUSION

This paper explores a fruits recognition classifier based on CNN algorithm. The accuracy and loss curves were generated by using various combinations of hidden layers for five cases using fruits-360 dataset. The recognition rate has greatly improved throughout the experiment. Among all the cases, the model achieved the best test accuracy of 100% in case 4 from 11 to 15 epochs and best training accuracy of 99.79% in case 1 at epoch 15. This type of higher accuracy will cooperate to stimulate the overall performance of the machine more adequately in fruits recognition. On the contrary, the highest and the lowest performance loss were found without and with the presence of dropout. The highest loss is approximately 0.5881 found in case 2 without dropout. Besides, the lowest performance loss is approximately 0.0032 in the presence of conv1, pool1, conv2, pool2 with dropout. This low loss will provide CNN better performance to attain better fruit recognition. In the future, our plan is to perform the segmentation process on the image before recognition and then applying it on CNN.